\documentclass[runningheads]{llncs}

\usepackage{eccv}

\usepackage{eccvabbrv}

\usepackage{graphicx}
\usepackage{booktabs}
\usepackage{multirow}

\usepackage[accsupp]{axessibility}  
\usepackage{hyperref}

\usepackage{orcidlink}

\begin{document}

\title{GroCo: Ground Constraint for Metric Self-Supervised Monocular Depth}


\author{Aurélien Cecille\inst{1,2}\and
  Stefan Duffner\inst{2}\orcidlink{0000-0003-0374-3814} \and
  Franck Davoine \inst{2}\orcidlink{0000-0002-8587-6997}\and
  Thibault Neveu\inst{1}\and
  Rémi Agier\inst{1}}

\authorrunning{A.~Cecille et al.}

\institute{Visual Behavior, Lyon, France \and
  INSA Lyon, CNRS, Ecole Centrale de Lyon, Universite Claude Bernard Lyon 1, Université Lumière Lyon 2, LIRIS, UMR5205, 69621 Villeurbanne, France}

\maketitle

\begin{abstract}
  Monocular depth estimation has greatly improved in the recent years but models predicting metric depth still struggle to generalize across diverse camera poses and datasets. While recent supervised methods mitigate this issue by leveraging ground prior information at inference, their adaptability to self-supervised settings is limited due to the additional challenge of scale recovery. Addressing this gap, we propose in this paper a novel constraint on ground areas designed specifically for the self-supervised paradigm. This mechanism not only allows to accurately recover the scale but also ensures coherence between the depth prediction and the ground prior. Experimental results show that our method surpasses existing scale recovery techniques on the KITTI benchmark and significantly enhances model generalization capabilities. This improvement can be observed by its more robust performance across diverse camera rotations and its adaptability in zero-shot conditions with previously unseen driving datasets such as DDAD.


\end{abstract}

\section{Introduction}
Depth estimation is a fundamental task in computer vision, offering crucial 3D insights for various applications such as robotics, augmented reality and intelligent vehicles. Specifically, in the realm of intelligent vehicles, accurate depth perception is vital for navigating safely by identifying and localizing potential obstacles.

Among the various methods, monocular depth estimation is particularly attractive due to its cost efficiency and broad availability across many systems. It presents a viable alternative to more expensive technologies like Lidar and stereo vision, promising wide applicability in real-world scenarios.
The advancement of robust monocular depth models, however, is hampered by the need for diverse, large-scale, annotated datasets, which are costly to create.\\
In response, there has been an increased interest in self-supervised learning, where models are trained using unlabeled data by leveraging the consistency of scene geometry across different viewpoints and moments in time.
Nonetheless, the self-supervision brings its own set of complications, notably in deriving metric scale information. This problem aligns with the longstanding challenges in the field of monocular visual odometry \cite{aqel2016review}, essential for self-supervised learning of depth. In fact, it is well-documented that the scale of the scene cannot be determined using only monocular images, introducing an inherent ambiguity in the predicted depth and egomotion, which are known only relative to an unknown scale factor. At its core, this issue arises because an image may correspond to various 3D scenes, differentiated only by their scale. This ambiguity is problematic as it obstructs applications that depend on precise distance measurements for decision-making.

\begin{figure}[tb]
  \centering
  \includegraphics[width=1.0\textwidth]{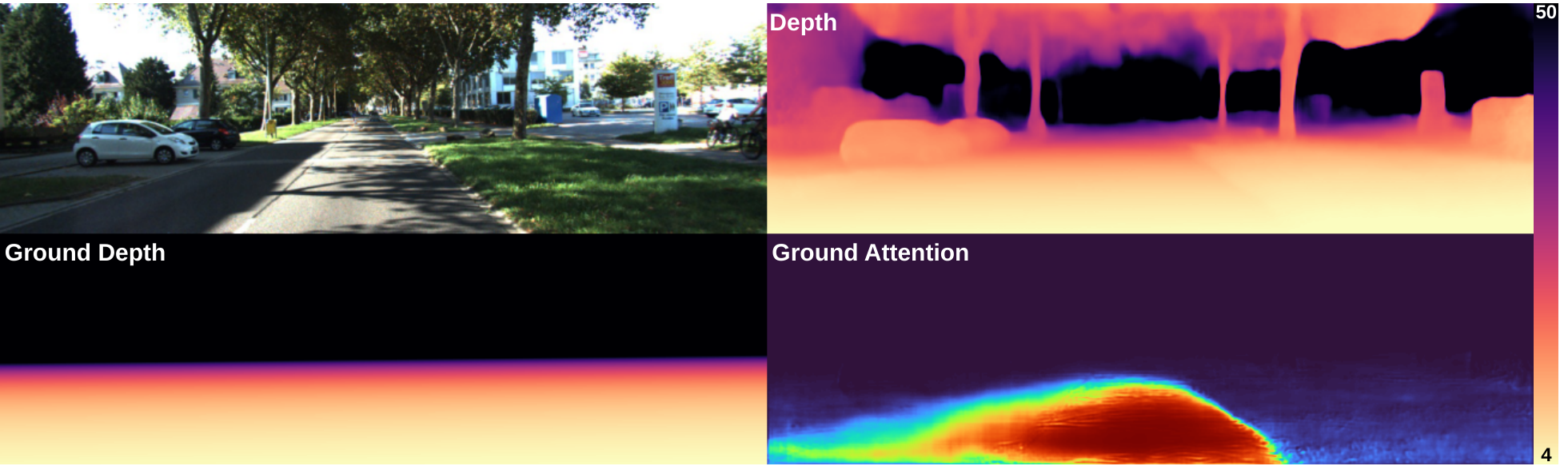}
  \caption{Example of the models' depth and ground attention prediction. The ground depth is given as input and integrated in the depth prediction using the attention map.}
  \label{fig:grid}
\end{figure}

Additionally, both self-supervised and supervised depth estimation methods often face the challenge of model overfitting to specific camera parameters \cite{van_dijk_how_2019}. Indeed, within the context of driving, monocular models often infer depth by correlating the vertical position of ground pixels with constant depths, a process heavily relying on unchanging camera intrinsic and extrinsic parameters. Such an approach significantly hampers the models' ability to generalize across different camera setups. Consequently, models require retraining or fine-tuning for each new camera configuration, which significantly limits their practical usability in diverse real-world environments.\\
In the context of ground-based systems like vehicles or robots, this limitation can be mitigated by using the theoretical flat ground as a reference since it can be deduced from commonly known camera parameters.
Two strategies for integrating this ground information exist: the \textit{a posteriori} method, which involves adjusting the scale of depth predictions during the inference phase \cite{xue2020toward}, requiring additional processing steps and necessitating ground segmentation; and the \textit{a priori} method, showcased in recent supervised learning approaches \cite{yang_gedepth_2023,koledic_gendepth_2023}, which embeds ground priors into the depth estimation model itself. This latter strategy equips the model with all necessary information for robust scaled depth prediction right from the start, aiming to enhance performance on all types of scenarios.

Despite these advancements, the transition to self-supervised settings remains impeded by the additional scale ambiguity challenge. This gap underscores the need for innovative methodologies capable of leveraging ground plane information effectively in the absence of explicit labels.\\
Our work directly addresses this challenge by introducing novel loss functions specifically designed for the integration of ground plane priors within a self-supervised learning framework as illustrated in \cref{fig:grid}. These innovations significantly enhance the depth estimation models' accuracy and generalizability, facilitating robust performance across diverse camera configurations and environments.\\
Our four key contributions are the following: \textbf{(1)}  A self-supervised method for metric depth estimation that enhances generalization across camera poses and datasets. \textbf{(2)} Novel loss functions for precise scale recovery. \textbf{(3)} A new way of integrating ground attention not requiring any depth annotation. \textbf{(4)} An interpretable attention mechanism to accurately localize flat ground areas in images.\\
The source code is available at \url{https://github.com/Visual-Behavior/GroCo}.

\section{Related Work}

\subsection{Monocular Self-Supervised Depth Estimation}


Self-Supervised Depth estimation is a task that has already been widely studied in the past few years. The main idea is to train a model to predict the depth of a scene without labels by exploiting its geometry. The most common way to do this, is to use the simultaneous learning of depth and egomotion \cite{zhou_unsupervised_2017,vijayanarasimhan_sfm-net_2017,guizilini_3d_2020,godard_digging_2019}. 
That is, the model is trained to predict the depth of the scene and the motion of the camera at the same time by computing the photometric error between the original image and the reprojected one from the predicted depth and motion. This is a very efficient way to train the model as it does not require any labels but uses the assumption of a static scene and a moving camera.
Godart \etal~\cite{godard_digging_2019} proposed a method that is robust when these hypotheses are not satisfied. They manage sequences where there is no egomotion by masking out pixels that do not change between frames and use a minimum loss across adjacent frames to handle dynamic objects.
\\
Model architectures have also been improved, Lyu \etal \cite{lyu_hr-depth_2021} enhanced the quality and sharpness of predicted depth, and recently, Transformer-based architectures have also been used to further increase performance \cite{zhang_lite-mono_2023,zhao_monovit_2022}.

\subsection{Scale Recovery}
Metric depth is crucial for downstream tasks. However, since images do not naturally reflect scale changes, incorporating additional information during training is essential for retrieving depth at the correct scale.\\
Guizilini \etal\cite{guizilini_3d_2020}, for example, presented a method leveraging vehicle velocity to impose a scale on egomotion estimation, constraining the depth estimation to be scaled as well.
Wagstaff and Kelly \cite{wagstaff_self-supervised_2021} proposed to scale the depth using the height of the camera. The process begins by training an up-to-scale model to derive relative depth. Following this, an unsupervised ground segmentation model is developed using the assumption that the bottom middle part of the image is the ground and fitting its relative depth to a plane. All pixels that are close to this plane are then considered as ground. In the subsequent stage, the scale is computed by fitting a plane on the segmented depth and scaling its normal vector with the camera height. New loss terms are then included in the optimization so that the model has to predict depth and egomotion that are equal to their scaled counterpart.
By exploiting "off-the-shelf" ground and vehicle segmentation models, Kinoshita and Nishino \cite{kinoshita_camera_2023-1} leverage the assumption of constant camera height to recover the scale of the scene. They especially use the fact that projecting vehicle points to the plane orthogonal to the ground always gives the same height even if the depth of objects changes. Combining this with the prior knowledge of the vehicle height allows to recover the camera height and the scale of the scene.
Zhang \etal \cite{zhang_towards_2022} recovers the scale using the IMU sensor combined with an extended Kalman filter (EKF) to provide motion at scale, constraining the depth to adopt the same metric scale.
Xiang \etal \cite{xiang_visual_2022-1} propose to recover the scale using the fact that in the KITTI dataset \cite{Uhrig2017THREEDV} the rectangular area in the middle bottom part of the image belongs to the ground. Combined with the camera height prior, it allows to determine the scale of the depth.\\
We notice that all methods that use the camera height rely on some form of ground segmentation, whether model or heuristic-based. This dependency introduces additional challenges in ensuring robustness across diverse scenarios, potentially restricting their usability.
\\
Additionally, most of these models consider the scale as a constant since they only use their prior during inference, and they do not generalize well when a change of camera position should result in scale adjustments.

\subsection{Ground Prior}
\begin{figure}[tb]
  \centering
  \includegraphics[width=1.0\textwidth]{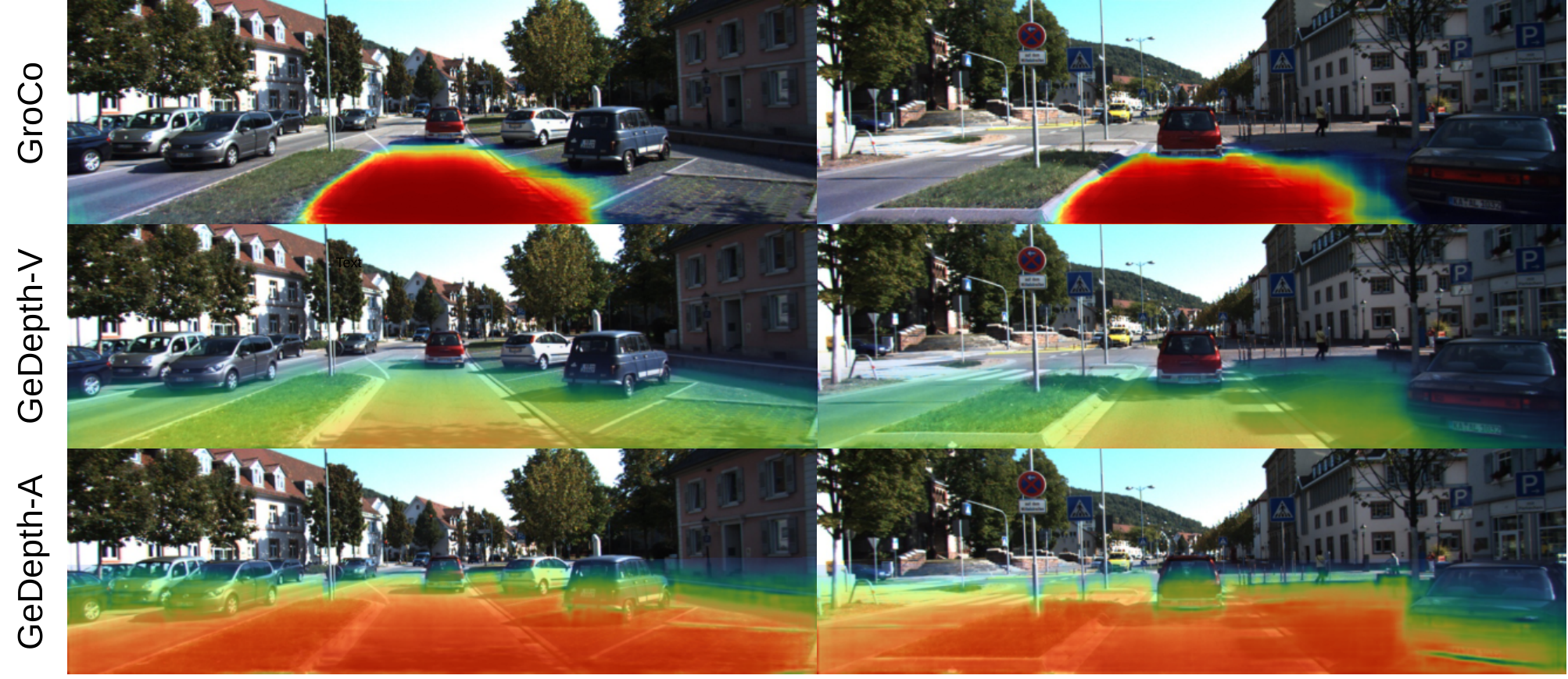}
  \caption{Result of attention maps compared to GeDepth \cite{yang_gedepth_2023}. While our method outputs very certain and precise ground segmentation, we see that GeDepth tends to have higher recall and uncertainty. We note that although Gedepth attention maps often consider the bottom part of obstacles as ground, it does not impact the end performance because these parts can be compensated by the residual depth or the slope prediction.  It also underlines the fact that their adaptive (A) version relies much more on the ground prior compared to the vanilla (V) one, potentially improving robustness.}
  \label{fig:attn}
\end{figure}

Van Dijk and De Croon \etal \cite{van_dijk_how_2019} demonstrated that monocular models estimate depth in two ways: by leveraging the vertical position of the contact point between the object and the ground and through the assimilation of a size prior for objects, with the former having a more significant impact.
They further highlighted the sensitivity of these methods to alterations in camera pose, leading to inaccuracies in ground recognition and consequently diminishing overall model efficacy.\\
To address this limitation, \cite{yang_gedepth_2023,koledic_gendepth_2023} have proposed the integration of a ground prior to provide camera pose information to the model and predict robust metric depth through the use of supervised annotations.\\
Koledić \etal \cite{koledic_gendepth_2023} employed a technique that transforms the ground plane into an embedding via a Fourier transform, which is then concatenated with encoder-derived features. This method trains on supervised synthetic data across a wide range of camera poses and thus exhibits robustness to these variations. It can be adapted to real data through a domain adaptation module and the utilization of stereo datasets.\\
The approach proposed by Yang \etal \cite{yang_gedepth_2023}, on the other hand,  normalizes the ground depth image and directly concatenates it with the input image. Additionally, the authors introduced a ground attention mechanism that works alongside the predicted depth to integrate the ground prior in the final output, termed \textit{Vanilla} version. Subsequently, they presented an \textit{Adaptive} version of their framework, capable of estimating the slope for each pixel within the ground prior, enhancing model accuracy in environments with uneven terrain.
Their findings suggest that this method not only generalizes more effectively to unseen datasets but also maintains robustness against changes in image resolution.
However, the slope estimation technique shows some limitations. In particular, its inability to adjust the horizon line restricts its applicability to merely offsetting existing ground pixels and consequently introduces issues with positive slopes. Besides, \cref{fig:attn} illustrates a counter intuitive behavior of the ground attention mechanism that considers the bottom part of obstacles as ground.\\
Despite the promise of these methodologies, they still require the use of stereo cameras or Lidar annotations to circumvent the scale issue inherent to monocular self-supervised settings — an aspect that our work addresses directly.

\section{Method}
This section outlines our methodology, demonstrating how each component synergistically contributes to solving the scale of the scene and improving generalization across diverse camera setups and datasets, thereby advancing the capabilities of self-supervised learning in depth estimation.
\subsection{Ground Plane}
\begin{figure}[tb]
  \centering
  \includegraphics[width=1.0\textwidth]{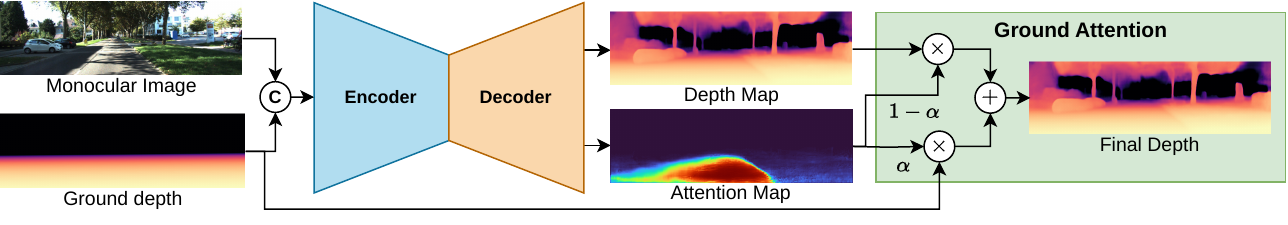}
  \caption{Illustration of the model architecture, highlighting the integration of ground depth information. The input image and ground depth are concatenated to provide ground aware features. The ground attention mechanism combines the depth map with the ground depth, guided by the attention map, to produce a refined final depth estimation.}
  \label{fig:archi}
\end{figure}
To provide the ground prior to the model we use the modifications proposed by \cite{yang_gedepth_2023} since the approach is very flexible and can be integrated with different types of neural network architectures such as CNN or Transformers.
\\
We compute the location of the theoretical ground plane thanks to the camera parameters and height $h$.
\\
Using the camera intrinsic $K =  \begin{bmatrix}
    f_x & 0   & c_x \\
    0   & f_y & c_y \\
    0   & 0   & 1
  \end{bmatrix}$ and extrinsic $E = \left[\begin{array}{c|c}
      R \  & \ t
    \end{array}\right]$ such that
\begin{equation}
  \begin{bmatrix}
    x \\
    y \\
    z
  \end{bmatrix}
  = R^{-1}(K^{-1}
  d_{u,v} \cdot
  \begin{bmatrix}
    u \\
    v \\
    1
  \end{bmatrix}
  -t)
\end{equation}
We can recover the depth of the ground $d_{u,v}$ for each pixels at position $(u, v)$ with height $y = h$ using the following formula:
\begin{equation}
  \label{plane_eq}
  d_{u,v} = \frac{h - t_y}{\frac{R_{1,2}}{f_x}(u-c_x) + \frac{R_{2,2}}{f_y}(v-c_y) + R_{3,2}}
\end{equation}
Computing the depth for all pixels and keeping only positive values, we obtain a ground depth image representing the distance from the camera to the theoretic ground at each pixel, ignoring obstacles and ground slope variations. This image is then normalized and directly concatenated with the input image as shown is \cref{fig:archi}.
\\

We adapt the ground attention scheme from \cite{yang_gedepth_2023}, utilizing its vanilla version.\\
The principle is to allow the model to choose between its own predicted depth $\hat{D}$ and the ground prior $G$ as can be seen in \cref{fig:archi}. It is done by adding a new attention map $\alpha$ such that the final depth for each pixel $i$ is obtained as follows:
\begin{equation}
  D_i = (1 - \alpha_i) \cdot \hat{D}_i + \alpha_i \cdot G_i
\end{equation}

\subsection{Scale Constraint}

\begin{figure}[tb]
  \centering
  \includegraphics[width=0.75\textwidth]{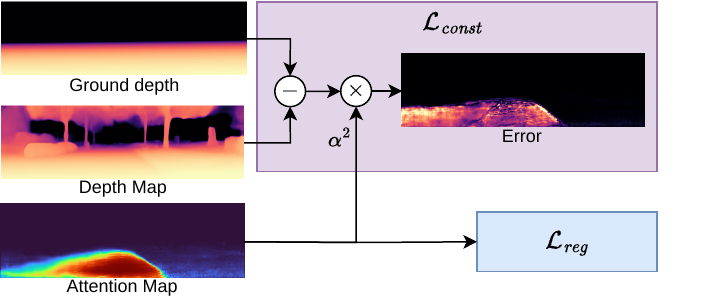}
  \caption{Overview of the proposed ground constraint loss $\mathcal{L}_{\mathit{const}}$ and attention regularisation $\mathcal{L}_{\mathit{reg}}$. The error image in $\mathcal{L}_{\mathit{const}}$ illustrates how this loss penalizes disagreement between the depth map and ground depth, indirectly ensuring that the scale of depth converges to the one of the ground.}
  \label{fig:loss}
\end{figure}

To ensure the accurate scaling of depth predictions, our approach incorporates two novel penalty terms during the training phase, as detailed in \cref{fig:loss}. These penalties are designed to more effectively leverage ground prior information, thereby guiding the model to implicitly estimate depth at the correct scale. The first penalty is an activation regularisation on the ground attention that ensures that it segments a minimum proportion of the ground. The second is a constraint loss that solves the scale issue and ensure that the attention correctly segments the ground area.\\

The regularisation addresses a fundamental challenge: without supervision, models do not automatically align the scale of their depth with the ground prior as they would when trained with labeled data, leading to the dismissal of the ground prior by the attention in favor of maintaining internal consistency in the depth estimates.\\
To counteract this tendency and promote the integration of the ground prior, we introduce a novel regularization term $\mathcal{L}_{\mathit{reg}}$. This term is designed to encourage the model to incorporate the ground prior into its depth estimation process by penalizing the attention when it does not activate enough, bridging the gap created by the absence of direct scale references from annotations.
However, since we do not want the model to take the ground depth everywhere, we only apply this regularisation up to a given threshold $\tau$ between 0 and 1, leading to the following formulation:
\begin{equation}
  \mathcal{L}_{\mathit{reg}} = \frac{\max(0, \tau - \frac{1}{N}\sum_i^N{\alpha_{i} })^2}{\tau^2}
\end{equation}
with $N$ the total number of pixels, $\alpha_i$ the $i^{th}$ pixel of the attention map and $\tau^2$ a normalization constant keeping the value in the unit interval.\\
We found that this formulation is much more robust to hyperparameters than using a classical regularisation while also being more intuitive to interpret.\\
Indeed, $\tau$ represents the proportion of the ground prior that we are confident at identifying as the ground, typically road surfaces. To prioritize precision and ensure the integrity and scale of depth estimations, it is recommended that $\tau$ be set below the proportion of the optimal ground segmentation.\\
Since $\tau$ changes depending on datasets, we propose a rule to compute its value based on the navigable area with respect to image dimensions $H$ and $W$ as well as the expected pathway width $P_w$ and camera height $h$: 
\begin{equation}
\small
     \tau = \frac{P_w H}{4hW} 
\end{equation}\\
Building on this, the constraint loss $\mathcal{L}_{\mathit{const}}$ ensures that the attention correctly activates on ground areas and that in these areas the predicted depth converges to the ground prior. The equality between the predicted depth and the ground prior is necessary so that the scale of the ground is correctly estimated and not degraded by the residual depth.
Its effect is twofold: it penalizes the attention on pixels where the ground prior and depth are distant, and, at the same time, penalizes the depth on pixels selected by the attention to make it converge to the ground prior. It can be expressed as:
\begin{equation}  \mathcal{L}_{\mathit{const}} = \frac{1}{N}\sum_i^N \alpha_i^2 |\hat{D}_i - G_i|
\end{equation}
We use an absolute distance instead of a relative one to ensure that the attention focuses on closer ground areas. These are more likely to meet the flatness criterion rather than distant areas where this assumption may not hold.
The attention is squared to penalize uncertain areas less and allow for a better quality depth prediction as opposed to the raw value that can cause the model to predict more binary attention maps.\\

We also use the reprojection loss  $\mathcal{L}_{\mathit{reproj}}$ and smoothness loss $\mathcal{L}_{\mathit{smooth}}$ from \cite{godard_digging_2019} to ensure that the model can estimate the geometry of the scene and correctly propagate the ground scale everywhere, resulting in the final loss:
\begin{equation}
  \mathcal{L} = \mathcal{L}_{\mathit{reproj}} + \lambda_{\mathit{smooth}}  \mathcal{L}_{\mathit{smooth}}  + \lambda_{\mathit{const}} \mathcal{L}_{\mathit{const}} + \lambda_{\mathit{reg}} \mathcal{L}_{\mathit{reg}} \; .
\end{equation}

Note that, compared to the adaptive method described in~\cite{yang_gedepth_2023}, we do not let the model predict the slope of the ground. This is for two reasons. The first is that if we give a new degree of freedom to the model, there would be no guarantee that it would converge to a metric depth. And the second is that it is not strictly necessary, since in case the ground is not flat, there is nothing stopping the model from simply discarding the area in the attention and predicting the correct depth.

\subsection{Rotation Augmentation}
\label{sec:rotation}
\begin{figure}[tb]
  \centering
  \includegraphics[width=1.0\textwidth]{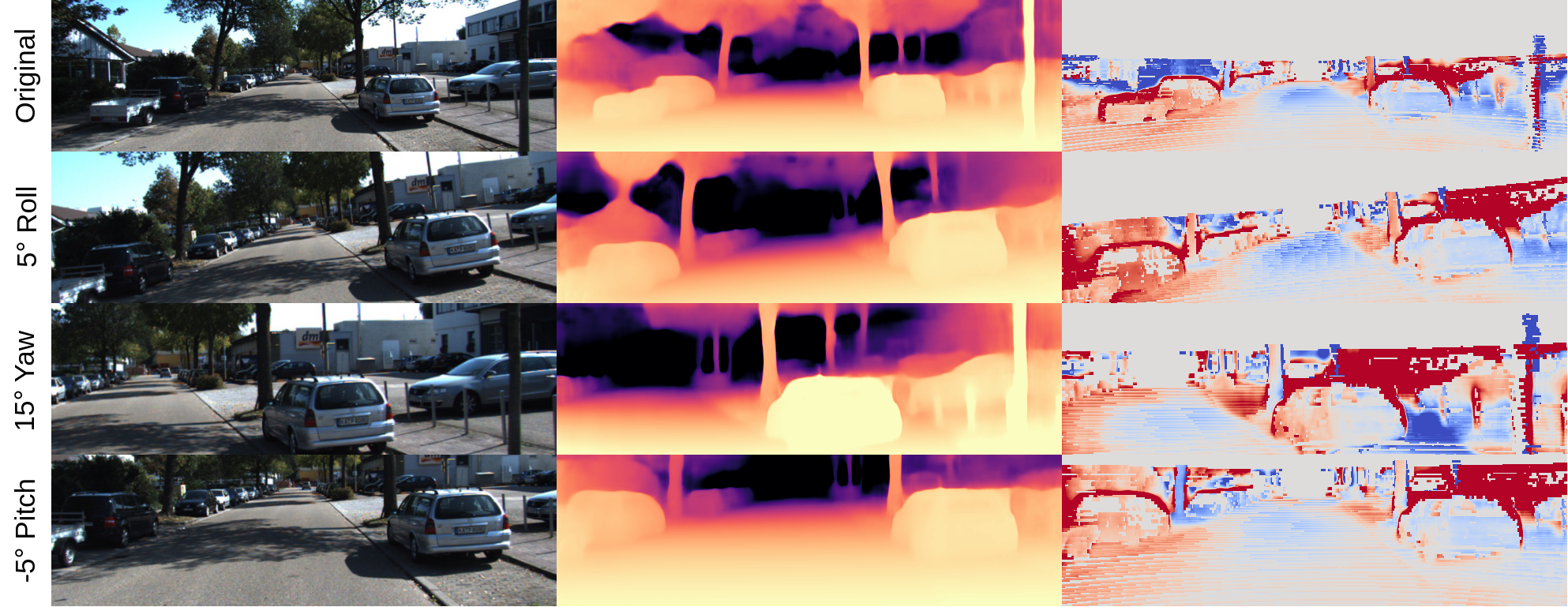}
  \caption{Visualisation of predictions on the same image with the various rotation augmentations. The last column is the relative per pixel error with the ground truth. The error is between -20\% in red and 20\% in blue, with 0\% or absence of ground truth in white.}
  \label{rot}
\end{figure}

To improve the robustness of the model, rotation augmentation is applied during training to both the images and coherently to the ground. This augmentation simulates camera pose changes and helps the model learn to handle different orientations of the scene. We focus on rotations since they can easily be simulated by warping images, compared to translations that would require to know a dense ground truth depth which is contradictory to the self-supervised setup.\\
We limit angles amplitudes to 5$^\circ$ for pitch and roll and 15$^\circ$ for yaw to not introduce any black borders or upscaling in images. The ground depth is also augmented to match the rotated images by directly applying the rotation on the camera extrinsic, avoiding interpolation errors. Illustrations of these augmentations can be seen in \cref{rot}.\\
In the same way, we also transform the Lidar depth used to evaluate the model to match the rotated images. This is done by rotating the Lidar point cloud and projecting it to the image plane to obtain the new depth.
\begin{figure}[b]
  \centering
  \includegraphics[width=\textwidth]{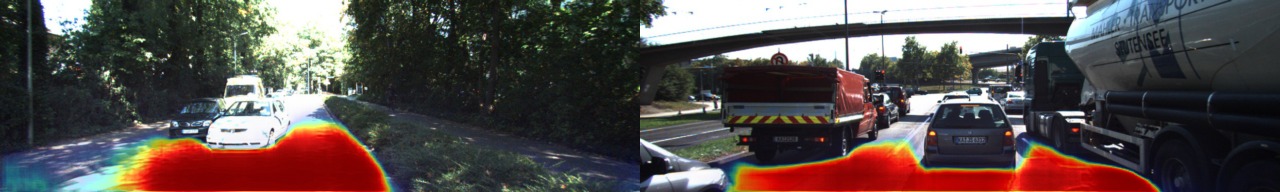}
  \caption{Example of segmentation quality of ground attention. Even in cases where there are close obstacles, the attention stays precise.}
  \label{attn_seg}
\end{figure}
\subsection{Interpretability}

Thanks to the ground attention mechanism, the prediction of the model can be reliably interpreted as seen in \cref{attn_seg}. By providing the area where the ground prior and the predicted depth are equal, we can detect failure cases of the model. This could typically be the case with images where the ground is not visible or if it is very uneven.
This information can be used either for humans to perform visual inspection or could for example be used during inference to filter predictions or trigger some sort of warnings in case the model is not able to detect any ground, potentially signaling that the camera has moved.

\section{Experiments}

\subsection{Implementation Details}
By default we follow \cite{godard_digging_2019} and use a Resnet50 encoder \cite{he2016deep} pretrained on the Imagenet dataset \cite{deng2009imagenet} and the same decoder coming from \cite{godard2017unsupervised}.
To take the additional inputs from the ground embedding channel, the pretrained weights are kept and the weights of the new  channel are initialized with a value of zero.
We adapt the outputs of the decoder, replacing the sigmoid by the softplus function to directly predict a strictly positive depth coherent with the ground prior. We also add a new head using the features at all resolutions to predict the attention map similarly to \cite{yang_gedepth_2023}. \\
For the hyper-parameters, we keep the default $\lambda_{smooth} = 10^{-2}$ and set $\lambda_{const} = \lambda_{reg} = 0.1$. $\tau$ is set to 0.25 on KITTI and 0.5 on DDAD to reflect the difference in image width and corresponds to a pathway width of two 2.75m wide lanes.\\
The model is trained using the Adam optimizer \cite{kingma2014adam} with a learning rate of $10^{-4}$ and a batch size of 12 for 20 epochs on KITTI \cite{geiger2013vision}. On an NVIDIA RTX 3090, it takes about 10 hours for the training to finish.\\

\subsection{Datasets}
We use the KITTI dataset extensively since it is the standard for depth estimation in the use case of intelligent vehicles. We report results using the eigen split using both the original Lidar data \cite{geiger2013vision} and the improved depth coming from the KITTI depth benchmark \cite{Uhrig2017THREEDV}. Unless specified, we will report results on the improved version since it more accurately represents the model performance.\\
We also use the DDAD dataset \cite{guizilini_3d_2020} to show the generalization of our model to new datasets and cameras. Similarly to \cite{guizilini2021sparse} we use the front, back, front left and front right cameras to evaluate the model.

\subsection{Performance}

\begin{table}[tb]
  \caption{KITTI self-supervised metric depth performance on two versions of labels. Comparison with methods that only use camera height to recover the scale. Results from other works are taken from \cite{kinoshita_camera_2023-1}.}
  \label{tab:KITTI_performance}
  \centering
  \centerline{
    \begin{tabular}{c|c|cccc|ccc}
      \toprule
      \multirow{2}{*}{Labels} & \multirow{2}{*}{Method}                            & \multicolumn{4}{c|}{Error ($\downarrow$) } & \multicolumn{3}{c}{Accuracy ($\uparrow$)}                                                                                                                        \\
                              &                                                    & AbsRel                                     & SqRel                                     & RMSE           & RMSE$_{\mathrm{log}}$ & $\delta\!<\!1.25$ & $\delta\!<\!1.25^2$ & $\delta\!<\!1.25^3$\vspace{0.5mm} \\
      \toprule
      \multirow{3}{1.5em}{\cite{geiger2013vision}}
                              & Scale Recovery\cite{wagstaff_self-supervised_2021} & 0.123                                      & 0.996                                     & 5.253          & 0.213                 & 0.840             & 0.947               & 0.978                             \\
                              & VADepth \cite{xiang_visual_2022-1}                 & 0.120                                      & 0.975                                     & 4.971          & 0.203                 & 0.867             & 0.956               & 0.979                             \\

                              & \textbf{Groco}                                     & \textbf{0.113}                             & \textbf{0.851}                            & \textbf{4.756} & \textbf{0.197}        & \textbf{0.870}    & \textbf{0.958}      & \textbf{0.980}                    \\
      \midrule
      \midrule
      \multirow{2}{2em}{\cite{Uhrig2017THREEDV}}
                              & VADepth  \cite{xiang_visual_2022-1}                & 0.091                                      & 0.555                                     & 3.871          & \textbf{0.134}        & \textbf{0.913}    & 0.983               & \textbf{0.995 }                   \\
                              & \textbf{Groco}                                     & \textbf{0.089}                             & \textbf{0.517}                            & \textbf{3.815} & \textbf{0.134}        & 0.910             & \textbf{0.984}      & \textbf{0.995}                    \\
      \bottomrule
    \end{tabular}
  }
\end{table}

\begin{figure}[b]
  \centering
  \includegraphics[width=\linewidth]{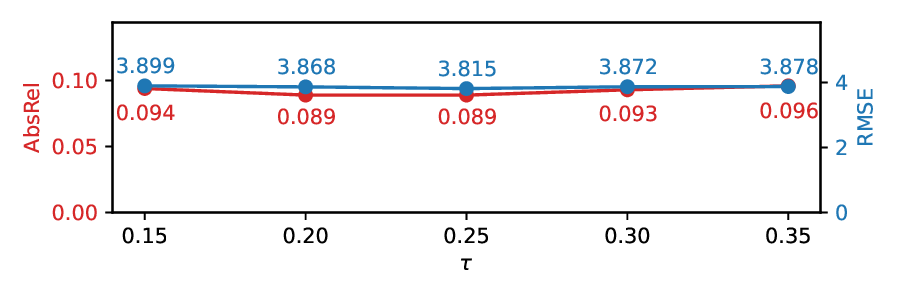}
   \caption{Model performance for varying $\tau$ on KITTI.}
   \label{fig:tau}
\end{figure}

We first compare our approach to the state-of-the-art methods that only use the camera height to recover the scale of the scene similarly to us.
We report standard metrics used for depth evaluation coming from \cite{eigen2014depth}: AbsRel (Absolute Relative Error), SqRel (Squared Relative Error) , RMSE (Root Mean Squared Error), RMSE$_{\mathrm{log}}$ (Root Mean Squared Log Error), $\delta < 1.25$, $\delta < 1.25^2$ and $\delta < 1.25^3$. $\delta$ metrics are accuracy measures and count the proportions of pixels where their ratio with the ground truth is inferior to $1.25^n$.
In \cref{tab:KITTI_performance} we see that our method outperforms the others in the monocular self-supervised metric depth estimation task without additional priors such as segmentation. \cref{fig:tau} shows the performance as a function of $\tau$ and its robustness with respect to it.

\subsection{Robustness to Camera Position Changes}
In order to evaluate our method against a comparable one, we propose a new baseline using the default Monodepth2 \cite{godard_digging_2019} pipeline in addition of losses proposed by \cite{wagstaff_self-supervised_2021} and leveraging the ground prior to estimate the scale, the method is detailed in the supplementary material. We compare the performance of both models on the KITTI dataset with different camera poses.
Both methods were trained with augmentation at training time. Results are reported in \cref{tab:rotation_performance}. We can see that our method performs better than the baseline for all rotations even though they perform very similarly on original images, demonstrating the gain of using our method to exploit the ground prior. For yaw and roll, we report positive values only since negative ones perform similarly. For the pitch we use the negative one because the positive augmentation leads to the ground not being visible in the image, rendering our method ineffective.\\
We also report the camera transfer performance against supervised methods in \cref{tab:camera_gen} and show that our method is able to better generalize to new cameras.

\begin{table}[tb]
  \caption{Performance on KITTI with different camera rotations.}
  \label{tab:rotation_performance}
  \centering
  \centerline{
    \begin{tabular}{c|c|cccc|ccc}
      \toprule

      \multirow{2}{*}{Augment}                       & \multirow{2}{*}{Method} & \multicolumn{4}{c|}{Error ($\downarrow$) } & \multicolumn{3}{c}{Accuracy ($\uparrow$)}                                                                                                                        \\
                                                     &                         & AbsRel                                     & SqRel                                     & RMSE           & RMSE$_{\mathrm{log}}$ & $\delta\!<\!1.25$ & $\delta\!<\!1.25^2$ & $\delta\!<\!1.25^3$\vspace{0.5mm} \\
      \toprule
      \multirow{2}{3em}{\textbf{None}}               & Baseline                & 0.092                                      & \textbf{0.499}                            & \textbf{3.701} & \textbf{0.134}        & \textbf{0.912}    & \textbf{0.984}      & \textbf{0.996}                    \\
                                                     & \textbf{Groco}          & \textbf{0.089}                             & 0.517                                     & 3.815          & \textbf{0.134}        & 0.910             & \textbf{0.984}      & 0.995                             \\
      \midrule
      \multirow{2}{3.5em}{\textbf{$5^{\circ}$ Roll}} & Baseline                & 0.140                                      & 0.879                                     & 5.083          & 0.194                 & 0.793             & 0.960               & 0.992                             \\
                                                     & \textbf{Groco}          & \textbf{0.101}                             & \textbf{0.635}                            & \textbf{4.301} & \textbf{0.147}        & \textbf{0.891}    & \textbf{0.979}      & \textbf{0.994}                    \\
      \midrule
      \multirow{2}{5em}{\textbf{$-5^{\circ}$ Pitch}} & Baseline                & 0.145                                      & 0.891                                     & 5.341          & 0.197                 & 0.782             & 0.952               & 0.989                             \\
                                                     & \textbf{Groco}          & \textbf{0.086}                             & \textbf{0.544}                              & \textbf{3.930} & \textbf{0.132}        & \textbf{0.914}    & \textbf{0.983}      & \textbf{0.995}                    \\
      \midrule
      \multirow{2}{3.5em}{\textbf{$5^{\circ}$ Yaw}}  & Baseline                & 0.107                                      & \textbf{0.548}                            & \textbf{3.944} & 0.149                 & 0.891             & 0.980               & \textbf{0.996}                    \\
                                                     & \textbf{Groco}          & \textbf{0.096}                             & 0.552                                     & \textbf{3.944} & \textbf{0.140}        & \textbf{0.900}    & \textbf{0.982}      & 0.995                             \\

      \midrule
      \multirow{2}{4em}{\textbf{$15^{\circ}$ Yaw}}   & Baseline                & 0.209                                      & 1.246                                     & 5.919          & 0.268                 & 0.483             & 0.932               & 0.986                             \\
                                                     & \textbf{Groco}          & \textbf{0.136}                             & \textbf{0.852}                            & \textbf{4.888} & \textbf{0.189}        & \textbf{0.808}    & \textbf{0.958}      & \textbf{0.989}                    \\
      \bottomrule
    \end{tabular}
  }
\end{table}
\begin{table}[tb]
  \centering
  \caption{Generalization on new cameras in the same domain compared to supervised methods. The model is trained on the front camera and evaluated on the other ones. Results from other works come from \cite{yang_gedepth_2023}, BTS \cite{lee2019big}  being a CNN-based architecture and DepthFormer \cite{li2022depthformer} a Transformer-based one.}
  \label{tab:camera_gen}
  \begin{tabular}{c|c|ccc}
    \toprule
    \multirow{2}{*}{Method}                                                           & \multicolumn{4}{c}{AbsRel ($\downarrow$)}                                                          \\
                                                                                      & Mean                                      & Back             & Left             & Right            \\
    \midrule
    DepthFormer \cite{li2022depthformer}                                              & 0.93                                      & 0.83             & 0.98             & 0.97             \\
    DepthFormer \cite{li2022depthformer}  + GeDepth Adaptive \cite{yang_gedepth_2023} & 0.66                                      & 0.64             & 0.59             & 0.75             \\
    BTS    \cite{lee2019big}                                                          & 0.72                                      & 0.82             & 0.98             & 0.97             \\
    BTS \cite{lee2019big}+ GeDepth Adaptive \cite{lee2019big, yang_gedepth_2023}      & \underline{0.62}                          & \underline{0.62} & \textbf{0.56}    & \textbf{0.67}    \\
    \midrule
    \textbf{Groco}                                                                    & \textbf{0.56}                             & \textbf{0.43}    & \underline{0.57} & \underline{0.68} \\
    \bottomrule
  \end{tabular}
\end{table}
\begin{table}[tb]
  \caption{Performance when trained on KITTI and evaluated on DDAD compared to supervised methods taken from \cite{yang_gedepth_2023}.}
  \label{tab:KITTI_to_ddad_mean}
  \centering
  \begin{tabular}{@{}c|cc|c@{}}
    \toprule
    Method                                    & AbsRel($\downarrow$) & RMSE ($\downarrow$)& Params \\ 
    \midrule
    DepthFormer \cite{li2022depthformer}      & 0.644                & 17.083             &  274M\\
    GeDepth-Adaptive \cite{yang_gedepth_2023} & \textbf{0.261}       & 16.132              & 277M\\
    \textbf{Groco}                            & 0.424                & \textbf{15.366}    & 35M\\
    \bottomrule
  \end{tabular}
\end{table}
\subsection{Generalization to New Datasets}
\begin{table}[tb]
  \caption{Performance when trained on KITTI and evaluated on DDAD for each camera.}
  \label{tab:KITTI_to_ddad_performance}
  \setlength{\tabcolsep}{2pt}
  \centering
  \centerline{
    \begin{tabular}{c|c|cccc|ccc}
      \toprule

      \multirow{2}{*}{Camera}            & \multirow{2}{*}{Method} & \multicolumn{4}{c|}{Error ($\downarrow$) } & \multicolumn{3}{c}{Accuracy ($\uparrow$)}                                                                                                                        \\
                                         &                         & AbsRel                                     & SqRel                                     & RMSE           & RMSE$_{\mathrm{log}}$ & $\delta\!<\!1.25$ & $\delta\!<\!1.25^2$ & $\delta\!<\!1.25^3$\vspace{0.5mm} \\
      \toprule
      \multirow{2}{3em}{\textbf{Front}}  & Baseline                & 0.403                                      & 5.366                                     & 14.364         & 0.567                 & 0.058             & 0.283               & 0.830                             \\
                                         & \textbf{Groco}          & \textbf{0.154}                             & \textbf{1.853}                            & \textbf{8.588} & \textbf{0.239}        & \textbf{0.760}    & \textbf{0.929}      & \textbf{0.975}                    \\
      \midrule
      \multirow{2}{2.5em}{\textbf{Back}} & Baseline                & 0.272                                      & 3.616                                     & 10.172         & 0.335                 & 0.586             & 0.856               & 0.937                             \\
                                         & \textbf{Groco}          & \textbf{0.233}                             & \textbf{3.031}                            & \textbf{9.753} & \textbf{ 0.318}       & \textbf{0.655}    & \textbf{0.874}      & \textbf{0.946}                    \\
      \midrule
      \multirow{2}{2em}{\textbf{Left}}   & Baseline                & 0.353                                      & 4.031                                     & 9.424          & 0.375                 & \textbf{0.798}    & \textbf{0.916}      & \textbf{0.975}                    \\
                                         & \textbf{Groco}          & \textbf{0.256}                             & \textbf{2.875}                            & \textbf{8.656} & \textbf{0.321}        & 0.647             & 0.861               & 0.937                             \\
      \midrule
      \multirow{2}{3em}{\textbf{Right}}  & Baseline                & 0.371                                      & 4.464                                     & 9.737          & 0.414                 & 0.466             & 0.752               & 0.883                             \\
                                         & \textbf{Groco}          & \textbf{0.334}                             & \textbf{3.832}                            & \textbf{9.190} & \textbf{ 0.389}       & \textbf{0.512}    & \textbf{0.790}      & \textbf{0.899 }                   \\

      \bottomrule
    \end{tabular}
  }
\end{table}
\begin{figure}[tb]
  \centering
  \includegraphics[width=1.0\textwidth]{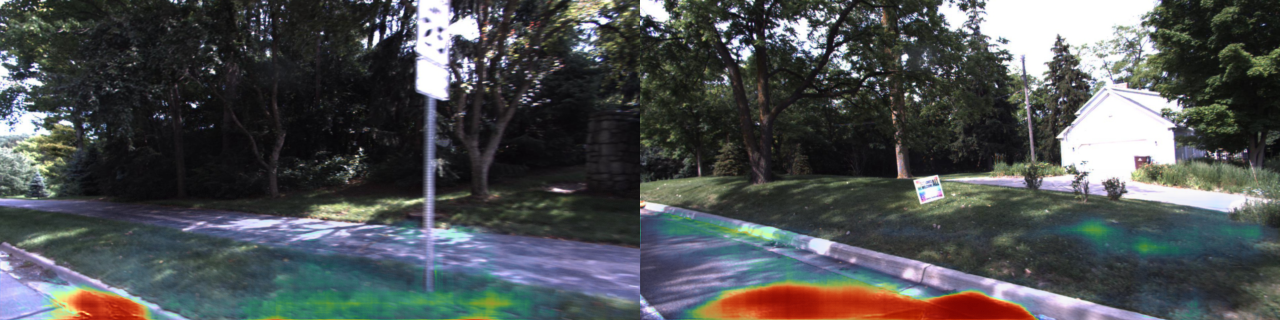}
  \caption{Example of the attention map for the side camera. Even on an unknown dataset with the road barely visible or sideways, the segmentation of the ground stays precise, ensuring good scale.}
  \label{side_attn}
\end{figure}

\begin{figure}[tb]
  \centering
  \includegraphics[width=\textwidth]{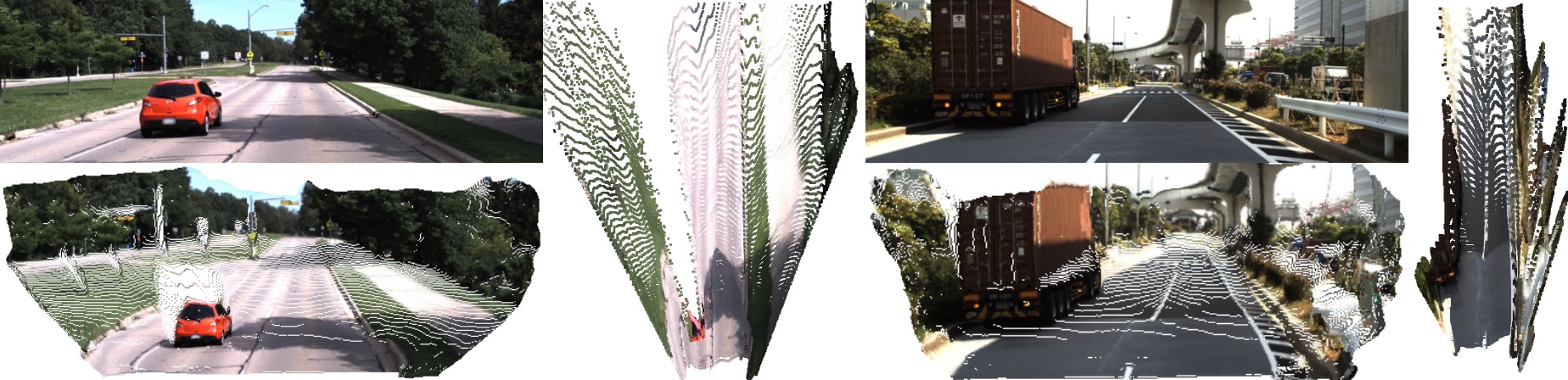}
  \caption{Point cloud reconstructions on the previously unseen DDAD dataset. The viewpoint of the bottom point cloud is translated 2m up and 8m back, while the side one is a top-down view. We can see that even though some artifacts are present, the overall shape and geometry of the scene is preserved.}
  \label{pc}
\end{figure}

We further evaluated the generalization capacity of our model by training it on the KITTI dataset and measuring its performance on the DDAD dataset. We report the results in \cref{tab:KITTI_to_ddad_performance}. These results are evaluated up to 80m and with an image height of 192 pixels like in the KITTI benchmark. We can see that our model generalizes better to the new dataset than the baseline for all cameras. We also notice that even if our model never saw images of side cameras, its attention is quite robust at segmenting the ground as can be seen in \cref{side_attn}.\\
\cref{tab:camera_gen} compares our method against the supervised results reported in \cite{yang_gedepth_2023}, using the same modalities as them. Point cloud reconstruction of our model are also demonstrated in \cref{pc}.
We see that despite having close to 8 times less parameters, the performance is quite similar to the supervised methods but vary depending on the metric used. We suspect that this gap comes from the fact that in DDAD the "ego-vehicule" is visible in images from the back and side cameras, potentially
impacting models performances differently.

\section{Limitations and Future Work}

Our approach is designed specifically for ground-based vehicles, leveraging the ground as a critical prior. This necessitates the ground's visibility within the field of view and presupposes the existence of at least a partially flat ground, which may limit its effectiveness on uneven terrains. This limit could be alleviated by propagating the scale across time to make sure that even if the flat ground is not visible for some time, the accuracy of depth can be conserved.\\
Additionally, our model depends on the parameter $\tau$, essential for successful training. Although \cref{side_attn} indicate that the model can adjust during inference to images with a smaller proportion of flat ground than $\tau$, the parameter stills needs to be set manually for each dataset in the training phase. \\
Future work could explore strategies to relax this constraint and enhance the ground attention mechanism's recall without sacrificing precision, which is vital for maintaining accurate scale estimation.

\section{Conclusion}

In this study, we introduced a novel self-supervised framework, GroCo, which enhances monocular depth estimation models by leveraging ground plane constraints to address scale ambiguity. Our approach significantly improves generalization across various camera setups and datasets, demonstrating comparable performance to supervised methods. By employing advanced loss functions that facilitate the incorporation of ground attention mechanisms without dependency on annotations, GroCo achieves significant advancements in scale recovery and metric depth estimation accuracy. These results highlight GroCo's potential in advancing the development of self-supervised learning frameworks for real-world applications.
\clearpage

\bibliographystyle{splncs04}
\bibliography{main}
\end{document}